\documentclass[conference]{IEEEtran}
\IEEEoverridecommandlockouts
\usepackage{cite}
\usepackage{amsmath,amssymb,amsfonts}
\usepackage{algorithmic}
\usepackage{graphicx}
\usepackage{textcomp}
\usepackage{xcolor}
\usepackage{bm}

\makeatletter
\renewcommand{\@cite}[2]{[#1\if@tempswa , #2\fi]}
\makeatother
\usepackage[hidelinks]{hyperref}
\usepackage{tikz}
\usepackage{booktabs}
\usepackage{pgfplots}
\usepackage{pgfplotstable}
\pgfplotsset{compat=1.18}

\def\BibTeX{{\rm B\kern-.05em{\sc i\kern-.025em b}\kern-.08em
    T\kern-.1667em\lower.7ex\hbox{E}\kern-.125emX}}
\begin{document}

\title{Verifying Rumors via Stance-Aware Structural Modeling
}
\iffalse
\author{\IEEEauthorblockN{WI-IAT'25 Anonymous Submission}
}
\fi

\author{\IEEEauthorblockN{Gibson\ Nkhata,\, Quan \ Mai,\, Uttamasha \ Anjally \ Oyshi,\,  Susan\ Gauch}
\IEEEauthorblockA{\textit{Department of Electrical Engineering and Computer Science} \\
\textit{University of Arkansas}\\
Fayetteville, AR 72701, USA \\
emails: \{gnkhata,\, quanmai,\, uoyshi,\, sgauch\}@uark.edu
}
}

\maketitle

\begin{abstract}
    Verifying rumors on social media is critical for mitigating the spread of false information. The stances of the conversation replies often provide important cues to determine a rumor’s veracity. However, existing models struggle to jointly capture semantic content, stance information, and conversation structure, especially under the sequence length constraints of transformer-based encoders. In this work, we propose a stance-aware structural modeling framework that encodes each post in a discourse with its stance signal and aggregates reply embeddings by stance category, enabling a scalable and semantically enriched representation of the entire thread. To enhance structural awareness, we introduce stance distribution and hierarchical depth as covariates, capturing stance imbalance and the influence of reply depth. Extensive experiments on benchmark datasets demonstrate that our approach significantly outperforms prior methods in the ability to predict the truthfulness of a rumor. We also demonstrate that our model is versatile for early detection and cross-platform generalization. %Our findings affirm the importance of integrating stance and structural signals for effective rumor verification.
\end{abstract}


\section{Introduction}\label{sec:intro}
Social media platforms play a pivotal role in information dissemination~\cite{ ozkent2022social}, yet they also foster the proliferation of rumors, which poses a significant challenge to society~\cite{kochkina2018all}.  A rumor is defined as a widely circulated piece of information whose veracity is uncertain~\cite{derczynski2017semeval}\cite{li2024filter}; it appears credible but lacks immediate verification and often provokes skepticism or concern, thus prompting users or automated systems to seek confirmation of its truthfulness, relying on machine learning and Natural Language Processing (NLP) techniques~\cite{yu2020coupled}\cite{luo2024joint}. 

Recent rumor verification models rely on three primary information sources: (1) the semantic content of the rumor post~\cite{derczynski2017semeval}, (2) retweet dynamics~\cite{naumzik2022detecting}, and (3) the content and structure of the resulting conversation thread~\cite{li2024filter}\cite{luo2024joint}. Large Language Models (LLMs), widely used for encoding post semantics due to their effectiveness, perform well in single-post settings. However, they struggle with encoding full rumor discourses because of sequence length constraints, limiting their ability to capture comprehensive semantic and stance-related cues in extended conversations.

Yu et al.~\cite{yu2020coupled} address long conversation threads by dividing them into shorter sub-threads, each encoded separately using BERT~\cite{devlin2019bertpretrainingdeepbidirectional}. A global layer then integrates these representations, subject to fixed limits on the number of posts and subthreads. While effective for short discussions, this method lacks scalability for longer threads. Moreover, splitting posts across subthreads compromises semantic integrity by fragmenting their context.

Li et al.~\cite{li2024filter} propose a stance-based neural model that verifies rumors by transforming stance sequences into frequency-domain controversy patterns, while leveraging conversation context and reply chains. Although each post is independently encoded using BERT, the model’s multi-stage stance extraction and frequency filtering introduce high computational overhead, potentially limiting its scalability. Additionally, its reliance on rigid thread structures (e.g., chronological order, consistent reply chains) reduces robustness to irregular or platform-specific conversation formats. A recent study~\cite{nkhata2025ipkm} tackles sequence length and platform-dependence challenges by embedding the rumor post separately with contextual embeddings and encoding stance labels using a Bidirectional Long Short Term (BiLSTM)~\cite{schuster1997bidirectional}. However, by decoupling stance labels from their corresponding text, the model fails to capture the underlying semantic-stance relationships effectively.

Rumor-aware stances are critical for effective rumor verification, yet remain difficult to exploit due to the short and context-limited nature of social media posts, which often hinders full semantic understanding~\cite{li2024filter}. To address these limitations, we propose a novel rumor verification model that captures both semantic content and stance dynamics within a rumor discourse. Each post in the thread is encoded using an LLM and augmented with its corresponding stance signal. Stance-specific aggregation of reply embeddings is then employed to mitigate sequence length constraints while preserving stance-related distinctions.

The number of replies to a post serves as a key indicator of rumor credibility~\cite{li2024samgat}. To retain this structural signal, we incorporate two structural covariates into our model: (1) Stance distribution, which quantifies the number of replies per stance category, offering insights into collective opinion and stance imbalance—both critical cues for rumor veracity~\cite{derczynski2017semeval}\cite{li2024filter}; and (2) hierarchical depth level, which captures the depth of each reply within the thread, aiding in modeling conversational influence. Usually, discourse posts closer to the source tend to be more relevant, while deeper-level replies often reflect rumor propagation and conversational drift through echo chambers or off-topic diversions~\cite{luo2024joint}.

To this end, we develop five Research Questions (RQs): \textbf{RQ1}: Can a stance-aware structural modeling approach enhance the rumor verification process? \textbf{RQ2:} Does our stance-aware structural model outperform state-of-the-art rumor verification methods across different settings? \textbf{RQ3:} How do the different constituents of the model contribute to the performance of the entire model? \textbf{RQ4:} Is our model effective for early rumor detection? and \textbf{RQ5:} Is our model suitable for cross-platform adaptation?

The primary contributions of this work are as follows:

\begin{itemize}
    \item  We propose a novel stance-aware structural modeling framework that encodes each post in a conversation thread using an LLM, while explicitly integrating stance information. 

    \item To overcome LLM sequence length constraints, we introduce a scalable stance-based aggregation method that compresses reply embeddings by stance category, preserving discourse-level stance structure regardless of thread length.

    \item We enhance structural modeling by incorporating stance distribution and hierarchical depth as covariates, capturing collective stance signals and conversational influence patterns critical for rumor verification.

    \item Extensive experiments on benchmark datasets demonstrate the effectiveness of our model, with ablation studies, early detection, and cross-platform adaptation confirming the importance of the proposed rumor model.
\end{itemize}

The remainder of this paper is organized as follows: Section~\ref{sec:rel} reviews related work on rumor verification; Section~\ref{sec:task} defines the task formulation; Section~\ref{sec:meth} details the proposed methodology; Section~\ref{sec:exp} presents experimental results; and Section~\ref{sec:conclusion} provides the conclusion.

\section{Related Work}\label{sec:rel}
Rumor verification research spans three main categories: content-based methods, structure-aware models, and approaches leveraging retweet dynamics.

\textbf{Content based methods}. Initial rumor verification methods primarily rely on textual content, using traditional machine learning and NLP techniques to extract handcrafted temporal and structural features~\iffalse\cite{ 10.1145/2350190.2350203}\fi\cite{ma2018detect}. With advances in deep learning, models such as LSTM have been introduced to automatically learn semantic patterns from rumor texts, often by organizing posts chronologically to represent thread-wide discourse~\iffalse\cite{lv2020rv}\fi\cite{luo2021bcmm,wang2022marv}. Despite their effectiveness, these approaches struggle to capture long-range dependencies and rely on sequential encoding. As a result, transformer-based models like BERT have become central to recent systems~\cite{luo2024joint}. However, such models are constrained by the maximum sequence length and often truncate critical replies, meanwhile overlooking the hierarchical structure of a discourse.

\textbf{Retweet dynamics methods.} Several statistical classifiers have leveraged retweet dynamics to verify rumors. For instance, Ducci et al.~\cite{ducci2020cascade} proposes a tree-structured framework that models information cascades through bidirectional encoding, capturing the propagation patterns of tweets and retweets. Their framework integrates retweet-derived features from inner nodes to enrich the global context of the cascade. Similarly, Naumzik et al.~\cite{naumzik2022detecting} introduces a probabilistic mixture model that distinguishes between true and false rumors based on retweet behaviors. While these studies use structural characteristics of retweet cascades, our work leverages stance dynamics of a discussion initiated by the source post in order to verify a rumor. 

%However, both approaches omit the discourse surrounding rumor posts.

\textbf{Structure-aware models.} Later studies have incorporated discourse structures to enhance model performance~\cite{ma2018tree,yuan2019jointly,wei2021towards}. Ma et al.~\cite{ma2018tree} have utilized tree-structured recursive neural networks to capture hierarchical reply dependencies, while graph-based models such as SAMGAT~\cite{bai2023samgat} and sequence graph networks~\cite{mai2024sequence} have leveraged attention mechanisms to model semantic and structural interactions in conversations. Temporal aspects of discourse have been explored by Sahni et al.~\cite{sahni2022temporal} and Luo et al.~\cite{luo2022dynamicgcn}. %, emphasizing the role of post timing in rumor propagation.
Additionally, several multitask learning frameworks have jointly addressed rumor verification and stance classification to exploit stance signals as auxiliary features~\cite{kochkina2018all}\cite{yu2020coupled,luo2024joint,li2024filter}\cite{ma2018detect}.

Deviating from existing work, this study proposes a single-task rumor verification framework centered on the discourse initiated by a source claim, prior to its diffusion through retweets. Formally, we extract content embeddings fused with stance signals and leverage stance-aware structural modeling in order to verify a rumor.

\section{Task Definition}\label{sec:task}
We define the rumor verification task over a set of rumor threads \( T = \{t_1, t_2, \dots, t_n\} \), where each thread \( t_i = (c, Y, R(c), S_i) \) represents a unique rumor event. Here, \( c \) denotes the source claim, \( Y\in\{\text{true (T)}, \text{false (F)}, \text{unverified (U)}\} \) is its veracity label, \( R(c) = \{r_1, r_2, \dots, r_m\} \) is the set of chronological replies, and \( S_i \) captures the structural properties of the thread (e.g., reply hierarchy). Each post \( v_i \in \{c, r_1, \dots, r_m\} \) is associated with a stance label \( s_i \in \{\text{support (S)} , \text{deny (D)}, \text{query (Q)}, \text{comment (C)}\} \). The objective is to learn a supervised classification function \( f: T \rightarrow Y \), which predicts the veracity of each thread \( t_i \) by jointly modeling its semantic and structural components.

\section{Methodology}\label{sec:meth}
\begin{figure}[t]
\centering
\includegraphics[width=0.45\textwidth]{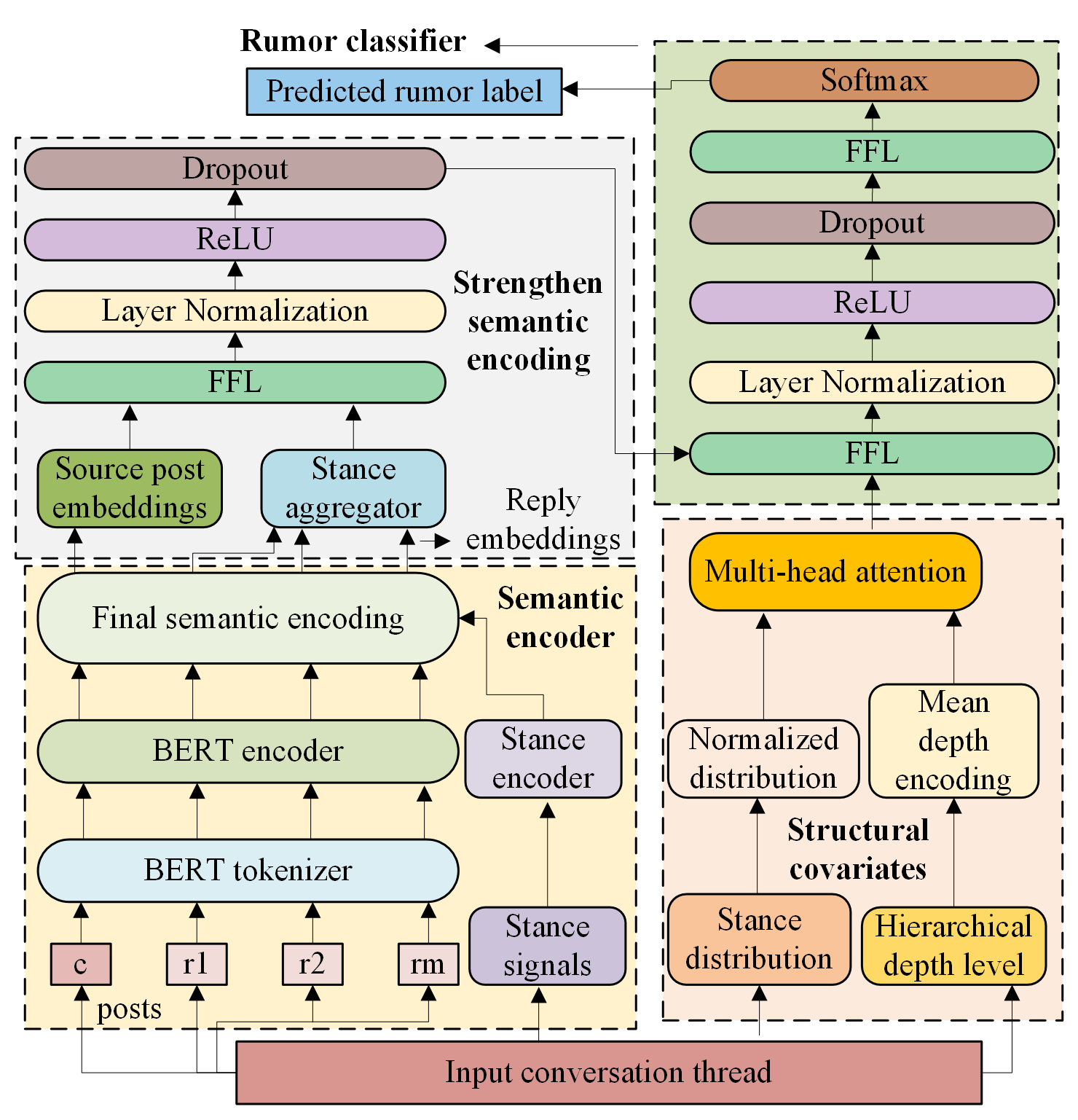} % Reduce the figure size so that it is slightly narrower than the column.
\caption{ The model framework, encompassing the semantic encoder, the stance aggregator, the structural covariate integrator, and the final rumor classifier.}
\label{fig:model}
\end{figure}

As shown in Fig.~\ref{fig:model}, the proposed model consists of four components: a \textbf{semantic encoder}, a \textbf{stance aggregator}, a \textbf{structural covariate integrator}, and a \textbf{rumor classifier}. The semantic encoder employs a pre-trained LLM to embed each post’s content along with its stance information. The stance aggregator compresses the variable-length reply thread into a structured, stance-aware representation. Finally, structural covariates are integrated with the semantic features and passed to the classifier for veracity prediction.

\subsection{Semantic Encoder}
Each post $v_i$ in a conversation thread is tokenized and passed through a pre-trained language model (e.g., BERT). This produces contextual embeddings $e(v_i)$ for each token in the post. A pooling strategy is then applied to aggregate these token embeddings into a single vector representing the entire post: $mean(e(v_i))$.      

On social media, users often express their stances in response to the position conveyed in the source of discussion. To enable the model to jointly represent the content and stance of each post, we inject explicit stance information into each post embedding by concatenating it with a one-hot encoded stance label vector as:

\begin{equation}
e'(v_i) = mean(e(v_i)) \mathbin\Vert s(v_i) \in \{0,1\}^4 ,
\label{eq:post_stanc}
\end{equation}

where $\mathbin\Vert$ denotes the vector concatenation operation, and integer 4 represents the stance categories. 

\subsection{Stance Aggregator}
 Social media posts are typically brief and provide limited contextual information~\cite{li2024filter}, making it difficult to verify a rumor based solely on the semantic content of the source claim. Since replies in a rumor discourse are annotated with stance labels, we note that the stance of each reply plays a crucial role in shaping the overall meaning of the conversation. Furthermore, replies sharing the same stance label often exhibit consistent linguistic patterns, such as similar framing of ideas and vocabulary choices. To capture this, we aggregate the embeddings of replies with the same stance to produce a single vector that summarizes the semantic contribution of that stance to the discourse, resulting in four aggregated stance embedding vectors. Let \( V_{\text{stance}} \subseteq V \) be the set of reply nodes in the thread that share a particular stance. The aggregated embedding for this stance is computed as:

\begin{equation}
e'(r_i) = \frac{1}{|V_{stance}|} \sum_{v_i \in V_{stance}} e'(v_i),
\label{eq:stance_vector}
\end{equation}

If no replies belong to a particular stance, a zero vector is used, $e'(r_i) = 0$. The stance aggregated reply embeddings $e'(r_i)$ are subsequently concatenated with the source claim encodings $e'(c)$ as

\begin{equation}
e'(t) = e'(c) \mathbin\Vert e'(r_1) \dots e'(r_4).
\label{eq:post_stanc}
\end{equation}
%Note that $e'(r)$ here encapsulates aggregated embeddings for all the four stance categories. 
Next, a Feed-Forward Layer (FFL) is applied to strengthen and learn a more abstract and discriminative semantic representation. Specifically, the concatenated vector $e'(t)$ is passed through a fully connected neural network composed of a linear transformation, layer normalization, a Rectified Linear Unit (ReLU) activation function, and a dropout layer for regularization and capturing nonlinear interactions between the semantic content and the stance information as

\begin{equation}
h(t) = Dropout(ReLU(LayerNorm(W_1 e'(t) + b_1))),
\label{eq:ffl}
\end{equation}

where $W_1$ and $b_1$ are the learned weight matrix and bias vector, respectively, and $h(t)$ is the resulting hidden representation for conversation thread $t$ after transformation. 

\subsection{Structural Covariate Integrator}

 Prior research~\cite{luo2024joint} has suggested that the structure of a discourse is essential for the model to understand the context to predict the veracity of a rumor. The stance aggregator solely concentrates on aggregation of the replies in the thread; however, to effectively capture the structural and semantic properties of discourse, we incorporate two auxiliary forms of information: (1) stance distribution and (2) hierarchical depth encoding. %and (iii) \textit{stance-wise semantic aggregation} of reply embeddings. These elements are combined with the source post representation to form a unified input to the model.

The stance distribution encodes the relative frequency of stance types among the replies to a claim post. Formally, let the stance distribution be represented as a mapping $ d: \mathcal{S} \rightarrow \mathbb{N},$
where \( \mathcal{S} = \{\text{S}, \text{D}, \text{Q}, \text{C}\} \) is the set of stance labels and $s_i$ is the frequency of a respective stance. The normalized stance distribution vector $ \mathbf{v}_d \in \mathbb{R}^{|\mathcal{S}|} $ is computed as
\begin{equation}
    v_d(R) = \frac{d(s_i)}{\sum_{j=1}^{|\mathcal{S}|} d(s_j)} \quad \text{for } s_i \in \mathcal{S}.
\label{eq:stanc_distr}
\end{equation}

%This vector \( v_d \) is used as a structural feature input alongside other representations. 
Afterwards, each reply post $r$ in the thread is assigned a hierarchical depth level, denoted \( h \in \mathbb{N} \), representing the number of reply hops from the source post. This depth is encoded as a one-hot vector \( h \in \{0,1\}^L \), where \( L \) is the maximum depth level allowed. Given that $i$ is an index in the vector, the encoding is defined as:
\begin{equation}
    d(r) = 
    \begin{cases}
        1 & \text{if } i = \min(h, L - 1) \\
        0 & \text{otherwise}
    \end{cases}.
\label{eq:hier_level}
\end{equation}
$\min(h, L - 1)$ ensures that even depth values exceeding the threshold are clipped and correctly encoded. To deal with scalability issues, where reply nodes in the thread increases, we compute the average hierarchy vector for replies of stance \( s \) following formulations of \eqref{eq:stance_vector} as

\begin{equation}
d'(r) = \frac{1}{|V_{stance}|} \sum_{r_i \in V_{stance}} d'(r_i).
\label{eq:hier_level_avg}
\end{equation}

The normalized \( v_d \) and the aggregated $d'(r)$ are combined and fed into a multihead attention (MHA)~\cite{vaswani2017attention} layer to enable the model to dynamically focus on pertinent parts of the structural covariates. For each attention head $i$, the scaled dot-product attention computes as

\begin{equation}
    head(i) =  Softmax\left(\frac{Q_i K_i^T}{\sqrt{d_k}}\right) V_i,
\label{eq:multiHead}
\end{equation}
where $Q_i$, $K_i$, $V_i$ are the query, key, and value matrices derived from input feature encodings through linear projections of $s' = v_d \mathbin\Vert d'(r)$ with weight matrices $W_i^Q$, $W_i^K$, and $W_i^V$as

\begin{equation}
Q_i = s' W_i^Q, \quad K_i = s' W_i^K, \quad V_i = s' W_i^V.
\end{equation}

The outputs from all heads are aggregated and passed through layer normalization and ReLU activation function, as in \eqref{eq:ffl}, resulting in $h_{att} = MultiHead(s')$.

\subsection{Rumor Classifier}

The final feed-forward classification module serves as the decision-making component that integrates both semantic and structural features to produce a probability distribution over rumor classes. Its input feature is computed as  

\begin{equation}
    z_{input} = [h(t) \; \| \; h_{att}] \in \mathbb{R}^{d_1 + d_2},
    \label{eq:final_input}
\end{equation}

where $d_1$ and $d_2$ are the dimensionalities of semantic features $h(t)$ from \eqref{eq:ffl}  and attended structural covariates $h_{att}$, respectively. These inputs are projected through two FFLs. The first layer maps the combined representation into a hidden space:

\begin{equation}
    h_{final} = Dropout \left( ReLU \left( LayerNorm \left( W_f z_{input} + b_f \right) \right) \right),
    \label{eq:ffl_final1}
\end{equation}

where $W_f$ and $b_f$ are learnable weights. The final output logits are computed, followed by a softmax operation to obtain a probability distribution over the rumor labels as

\begin{equation}
    \hat{\mathbf{y}} = Softmax(W_o h_{final} + b_o) \in \mathbb{R}^{C},
    \label{eq:final_ffl2}
\end{equation}

where $W_o$ and $b_o$ are trainable parameters and $C$ is the number of rumor classes. 

\subsection{Objective Function}
Due to the uneven distribution of rumor classes in the datasets employed in our experimentation, we design the regularization cost function for the whole framework by adapting the Focal Loss (FL)~\cite{Lin_2017_ICCV}. This loss function addresses class imbalance by dynamically reshaping the standard cross-entropy loss such that it down-weights the loss assigned to accurately classified samples and focuses learning on hard, misclassified examples. The standard cross-entropy loss is computed as 

\begin{equation}
    \mathcal{L}_{\text{CE}} = -\sum_{i=1}^C y_i \log(\hat{y}_i) = -\log(\hat{\mathbf{y}}),
    \label{eq:ce_loss}
\end{equation}
where $y$ denotes gold rumor labels and $\hat{\mathbf{y}}$ is computed from \eqref{eq:final_ffl2}. FL extends \eqref{eq:ce_loss}  by adding a modulating factor $(1 - \hat{\mathbf{y}})^\gamma$ and an optional class weighting factor $\alpha_{\hat{\mathbf{y}}} \in \mathbb{R}^C$:

\begin{equation}
\mathcal{L}_{\text{Focal}} = -\alpha_{\hat{\mathbf{y}}} (1 - \hat{\mathbf{y}})^\gamma \log(\hat{\mathbf{y}}), 
\end{equation}

%New equation

\begin{equation}
\mathcal{L}_{\text{Focal}} =
     -\sum_{i=1}^C \alpha_{yi} (1 - y_i)^\gamma \log(y_i),
    \label{eq:foc_loss}
\end{equation}

where $\gamma \geq 0$ is the focusing parameter that adjusts the rate at which easy examples are down-weighted. When $\gamma = 0$, the FL reduces to the cross-entropy loss. The factor $(1 - \hat{\mathbf{y}})^\gamma$ ensures that the contribution of well-classified examples (i.e., those with high $\hat{\mathbf{y}}$) is diminished, thus placing more emphasis on hard, misclassified examples. The loss is finally aggregated using mean reduction strategy. We compute $\alpha_{\hat{\mathbf{y}}}$ using a custom inverse frequency mechanism. Let $N$ denote the total number of training samples, and let $n_i$ represent the number of samples belonging to class $i$, where $i \in \{0, 1, 2\}$ for $\{\text{T},\text{F},\text{U}\}$. The weight assigned to class $i$ is given by $w_i = \frac{N}{n_i}$ if $n_i > 0$, and $w_i = 0$ otherwise. This formulation ensures that underrepresented classes are assigned higher weights during training, thereby emphasizing their contribution to the loss function. 

\section{Experiments}\label{sec:exp}
%We evaluate the performance of our proposed model against state-of-the-art and competitive baselines on publicly available datasets.
The experiments reported in this section aim to answer the research questions: \textbf{RQ1}, \textbf{RQ2}, \textbf{RQ3}, \textbf{RQ4}, and \textbf{RQ5}.

\subsection{Datasets}
We evaluate our model on three publicly available rumor verification benchmark datasets: RumEval2017~\cite{derczynski2017semeval}, RumEval2019~\cite{gorrell-etal-2019-semeval}, and PHEME~\cite{zubiaga2016learning}. RumEval2019 and PHEME are the extensions of RumEval2017, which consists of 325 rumor events and 5,568 tweets related to eight major breaking news stories on Twitter (now X). Each source claim is annotated with a veracity label, and each post in the thread is labeled with a stance category.

RumEval2019 extends RumEval2017 by incorporating additional test tweets and Reddit data, while reusing RumEval2017 events for training. It comprises 446 rumor threads and 8,574 posts, annotated with both veracity and stance labels.

The PHEME dataset expands RumEval2017 by including data from nine breaking news events, totaling 2,402 threads and 105,354 tweets. Unlike RumEval2019, the additional posts in PHEME are annotated only with veracity labels. 

Descriptive statistics of the datasets are presented in Tables~\ref{tab:RumEval17_stats}, \ref{tab:rumEval19_stats}, and  \ref{tab:pheme_stats}. In the tables, \textit{AvgDepth} stands for average hierarchy depth and \textit{NS} refers to posts without stance labels. RumEval2017 and RumEval2019 contain the original train/validation/test splits, whereas PHEME does not include a predefined partition.

\renewcommand{\arraystretch}{1.2}
\begin{table}[t]  
\centering
\caption{Descriptive statistics of RumEval2017} 
%\resizebox{0.9\textwidth}{!}{
\resizebox{1.0\linewidth}{!}{%  
\begin{tabular}{lcccccccccc}  
\hline  
\textbf{Split}  & \multicolumn{5}{c}{Rumor Statistics} & & \multicolumn{4}{c}{Stance Distribution} \\  
\cline{2-6} \cline{8-11}   
 \textbf{} & \textbf{Threads} & \textbf{AvgDepth} & \textbf{F} & \textbf{T} & \textbf{U} & & \textbf{S} & \textbf{D} & \textbf{Q} & \textbf{C} \\
\hline
Train set         & 272 & 3.2 & 50 & 127 & 95  & & 841 & 333 & 330 & 2734 \\
Development set   & 25  & 3.4 & 12 & 10  & 3    & & 69  & 11  & 28  & 173  \\
Test set          & 28  & 2.8 & 12 & 8   & 8   & & 94  & 71  & 106 & 778  \\
\hline
\textbf{Total}    & \textbf{325 }& \textbf{3.2} & \textbf{74} & \textbf{145} & \textbf{106}  & & \textbf{1004} &\textbf{ 415} & \textbf{464} & \textbf{3685} \\
\hline
\end{tabular}%  
}  
 
\label{tab:RumEval17_stats} 
\end{table} 

%Another Table
\renewcommand{\arraystretch}{1.2}
\begin{table}[h]  
\centering
\caption{Descriptive statistics of RumEval2019} 
%\resizebox{0.9\textwidth}{!}{
\resizebox{1.0\linewidth}{!}{%  
\begin{tabular}{lcccccccccc}  
\hline  
\textbf{Split}  & \multicolumn{5}{c}{Rumor Statistics} & & \multicolumn{4}{c}{Stance Distribution} \\  
\cline{2-6} \cline{8-11}   
 \textbf{} & \textbf{Threads} & \textbf{AvgDepth} & \textbf{F} & \textbf{T} & \textbf{U} & & \textbf{S} & \textbf{D} & \textbf{Q} & \textbf{C} \\
\hline
Twitter Train         &325 &2.2  &74  &145   &106   & &1004  &415  &464  &3685  \\
Reddit Train          &40  &3.0  &24  &9     &7     & &23    &45   &51   &1025   \\
Total Train           &365 &2.3  &98  &154   &113   & &1027  &460  &515  &4700   \\
\hline
Twitter Test          &56  &0.9  &30  &22    &4     & &141   &92   &62   &771 \\
Reddit Test           &25  &2.9  &10  &9     &6     & &16    &54   &31   &705   \\
Total Test            &81  &1.7  &40  &31    &10    & &157   &146  &93   &1476   \\
\hline
\textbf{Total}    & \textbf{446 }& \textbf{2.3} & \textbf{138} & \textbf{185} & \textbf{123}  & & \textbf{1184} &\textbf{ 606} & \textbf{608} & \textbf{6176} \\
\hline
\end{tabular}%  
}  
 
\label{tab:rumEval19_stats} 
\end{table} 

%Another Tabble
\renewcommand{\arraystretch}{1.2}
\begin{table}[h] 
\centering
\caption{Descriptive statistics of PHEME} 
\resizebox{1.0\linewidth}{!}{%  
\begin{tabular}{lccccccccccc}  
\hline  
\textbf{Event}  & \multicolumn{5}{c}{Rumor Statistics} & & \multicolumn{5}{c}{Stance Distribution} \\  
\cline{2-6} \cline{8-12}   
 \textbf{} & \textbf{Threads} & \textbf{AvgDepth} & \textbf{F} & \textbf{T} & \textbf{U} &  & \textbf{S} & \textbf{D} & \textbf{Q} & \textbf{C} & \textbf{NS} \\
\hline
Charlie Hebdo        & 458 & 3.5 & 116 & 193 & 149 &  & 248 & 60  & 61  & 795 & 380 \\
Sydney siege         & 522 & 3.3 & 86  & 382 & 54  &  & 225 & 90  & 110 & 769 & 448 \\
Ferguson             & 284 & 5.2 & 8   & 10  & 266 &  & 191 & 95  & 116 & 784 & 234 \\
Ottawa shooting      & 470 & 2.9 & 72  & 329 & 69  &  & 171 & 78  & 83  & 568 & 407 \\
Germanwings-crash    & 238 & 3.1 & 111 & 94  & 33  &  & 80  & 16  & 43  & 244 & 209 \\
Putin missing        & 126 & 2.2 & 9   & 0   & 117 &  & 18  & 6   & 5   & 33  & 117 \\
Prince Toronto       & 229 & 2.3 & 222 & 0   & 7   &  & 21  & 7   & 11  & 64  & 217 \\
Gurlitt              & 61  & 1.3 & 0   & 59  & 2   &  & 0   & 0   & 0   & 0   & 61  \\
Ebola Essien         & 14  & 2.4 & 14  & 0   & 0   &  & 6   & 6   & 1   & 21  & 12  \\
\hline
\textbf{Total}       & \textbf{2402} & \textbf{3.6} & \textbf{638} & \textbf{1067} & \textbf{697} &  & \textbf{960} & \textbf{358} & \textbf{430} & \textbf{3278} & \textbf{2085} \\
\hline
\end{tabular}%  
}  
\label{tab:pheme_stats}  
\end{table}

\subsection{Data Preprocessing}\label{Exp_data_prep}
In addition to standard data preprocessing steps, e.g., removing null entries, we apply several text normalization techniques. Hashtags (e.g., \#StopTheRumors100Times) are processed by stripping the leading \# and segmenting them into constituent words and numbers using pattern-based rules, yielding sequences like ``Stop The Rumors 100 Times.'' All hyperlinks are replaced with \textit{\$url\$}, and user mentions are replaced with \textit{\$mention\$}, following the protocol in~\cite{luo2024joint}. Emojis, which often carry sentiment or contextual cues, are converted into textual descriptions using the Python emoji library.

\subsection{Experimental Setup}

We adhere to the train/validation/test splits specified in the original papers for RumEval2017 and RumEval2019. Since PHEME lacks an official dataset division, we follow the conventional evaluation strategy for this corpus from previous works~\cite{kochkina2018all}\cite{li2024filter,yu2020coupled,luo2024joint}  by performing the k-fold cross-validation. For clarity, we implement the leave-one-event-out setting, testing each event individually and averaging the results from nine folds to compute the final performance.

Motivated by its success in recent NLP studies~\cite{li2024filter}\cite{luo2024joint}\cite{Nkhata2024Sarcasm}, we leverage BERT~\cite{devlin2019bertpretrainingdeepbidirectional} implemented by Huggingface~\cite {wolf2020transformers} to generate word embeddings for each post in our experiments. We simultaneously test an alternative pre-trained language model (PLM), a Robustly Optimized BERT Approach (RoBERTa)~\cite{liu2019roberta}, but it yields suboptimal performance compared to BERT, so it is dropped. 

During training, the model processes 16 rumor events per batch. The BERT tokenizer is configured with a maximum sequence length of 20, the average post length in all datasets. The model is trained for a total of $100$ epochs with a learning rate $\eta = 3 \times 10^{-4}$, weight decay coefficient $\lambda = 5 \times 10^{-6}$, and an optional gradient clipping threshold of $1.0$ to mitigate exploding gradients during backpropagation. Depending on the specific dataset being used, the best model checkpoint is saved to disk for later testing. The Adam optimizer~\cite{kingma2014adam} is employed to update the model parameters. Other hyperparameters include a dropout probability of 0.5 and four attention heads.  We meticulously experiment with various hyperparameters using the validation datasets before committing to these optimals, and thus present the tuning grid for hyperparameter optimization in Table~\ref{tab:hyperparam_tune_grid}.

\iffalse
\begin{itemize}
    \item Learning rate 	 $3 \times 10^{-4}$, $3 \times 10^{-3}$, $1 \times 10^{-3}$, $1 \times 10^{-4}$ 
    \item Weight decay 	$5 \times 10^{-6}$ 
    \item Dropout probability	 0.2, 0.25, 0.3, 0.4, 0.5 
    \item Batch size 	8, 16, 32, 64
    \item Epochs 	10, 30, 50, 70, 100
    \item Att. heads   1, 2, 4, 6, 8, 10 
    \item Max seq length  20, 32, 64, 128, 256, 512 
    \item Dep\_embed dimension 1, 3, 24
    \item FL gamma \&	 1, 1.5, 1.8, 2
\end{itemize}
\fi

Note that PHEME contains only partial stance annotations. To enrich these annotations, we first identify overlapping conversation threads between PHEME and RumEval2017 and extend them using the stance labels from RumEval2017. For the remaining unlabeled posts, we train an initial model (excluding the stance-based embedding aggregation and stance distribution modules) on the combined stance-labeled RumEval2017 and RumEval2019 datasets. Given the heavy class imbalance, particularly towards the comment stance, we apply the SMOTE oversampling technique~\cite{chawla2002smote} to balance the stance distribution and enhance model generalization. The best-performing model from this phase is then used to predict stance labels for the remaining PHEME posts. Importantly, SMOTE is not applied during the final rumor verification stage to ensure fair comparison with baseline methods.

\begin{table}[t]  
\centering
\caption{Hyperparameters Tuning Grid} 
%\resizebox{0.9\textwidth}{!}{
\resizebox{0.80\linewidth}{!}{%  
\begin{tabular}{lc}  
\hline  
\textbf{Hyperparameter} &  \textbf{Tested values}   \\  
\hline 

Learning rate &	 $3 \times 10^{-4}$, $3 \times 10^{-3}$, $1 \times 10^{-3}$, $1 \times 10^{-4}$ \\
Weight decay &	$5 \times 10^{-6}$ \\
Dropout  &	 0.2, 0.25, 0.3, 0.4, 0.5 \\
Batch size &	8, 16, 32, 64\\
Epochs &	10, 30, 50, 70, 100\\
Att. heads  & 1, 2, 4, 6, 8, 10 \\
FFL hidden size & 8, 16, 24, 32, 64, 128\\ 
Max seq length & 20, 32, 64, 128, 256, 512 \\
Dep\_embed dim.  & 1, 3, 24\\
FL gamma &	 0, 1, 1.5, 1.8, 2\\
FL reduction  & mean, sum\\

\hline
\end{tabular}%  
}  
 
\label{tab:hyperparam_tune_grid} 
\end{table}

\subsection{Model Evaluation}
Given the class imbalance across all datasets used in this study, relying solely on accuracy evaluation metric is insufficient, as models may achieve high accuracy by favoring the majority class~\cite{li2024filter}\iffalse\cite{lukasik2019gaussian}\cite{zubiaga2016stance}\fi. Following prior work~\cite{derczynski2017semeval, li2024filter, luo2024joint, yu2020coupled}, we evaluate model performance using both accuracy and Macro-F1, with the best model on the development set selected for final testing on RumEval2017 and RumEval2019. To further mitigate class imbalance, we compute dynamic class weights based on label frequencies and integrate them into the regularization loss function during training. All experiments are conducted on two Quadro RTX 8000 GPUs, 48 GB VRAM each.

\subsection{Baseline Models}
 The performance of the proposed model is evaluated against multiple state-of-the-art rumor verification baselines. While most of these baselines also leverage structural dynamics, we categorize them into two groups based on whether stance signals are utilized in the respective model.

\subsubsection{Baselines Excluding Stances}
\begin{itemize}
    \item \textbf{eventAI}~\cite{li-etal-2019-eventai}: Securing the first position in the RumEval2019 competition task~\cite{gorrell-etal-2019-semeval}, it leverages multidimensional information and employs an ensemble strategy to enhance rumor verification.

    %\item \textbf{HiTPLAN}~\cite{khoo2020interpretable}: It stacks two randomly initialized Transformer encoders as a hierarchical system, where the bottom Transformer encoder obtains the sentence representation of each tweet by aggregating its token-level information. After that, the post-level aggregation is used for rumor verification by the upper Transformer encoder.

    \item \textbf{SAMGAT}~\cite{li2024samgat}:  
    This model utilizes Graph Attention Networks (GATs)  to  capture contextual relationships between posts. While initially applied to the PHEME dataset for binary classification (excluding the \textit{Unverified} class), we adapt SAMGAT for our experimental setup. 

    \item \textbf{Knowledge Graphs (KGs)}~\cite{dougrez2024knowledge}:  
    This study proposes a knowledge graph-based methodology that automatically retrieves evidence for rumor verification.

    \item \textbf{Hierarchical Transformer (HT)}~\cite{yu2020coupled}:  
    This approach segments rumor threads into multiple groups based on the hierarchical structure of conversations. %Each group is processed using BERT to extract contextual information, and the aggregated information is fused using a Transformer for rumor verification. 
    
    \item \textbf{FSNet w/o stance}~\cite{li2024filter}: decomposes the reply sequence in the time domain into multiple controversies in the frequency domain and obtains a weighted aggregation of them.
    
\end{itemize}

\subsubsection{Baselines Including Stances}
\begin{itemize}
     \item \textbf{Coupled Hierarchical Transformer (CHT)}~\cite{yu2020coupled}:  
    Building on the HTs, it further enhances performance by integrating stance information. 

     \item \textbf{MTL-SMI}~\cite{liu2022predicting}: It comprises two shared channels and two task-specific graph channels. The graph channels are used to enhance the task-specific structure features for integrating rumor verification and stance classification.
    
    \item \textbf{Joint Rumor and Stance Model (JRSM)}~\cite{luo2024joint}: utilizes a graph transformer to encode input data and a partition filter network to explicitly model rumor-specific, stance-specific, and shared interactive features for rumor verification.  

    \item \textbf{FSNet with stance}~\cite{li2024filter}: An extension of  FSNet w/o stance. It supervises the stance extractor from reply sequencing with stance labels.

     \item \textbf{S-Conditioned Modeling (S-CoM)}~\cite{nkhata2025ipkm}: This work aggregates post embeddings and separately models stance progression with a BiLSTM for rumor verification.
        
\end{itemize}

\subsection{Results and Discussion} 
\subsubsection{\textbf{Comparison Results} (\textbf{RQ1} and \textbf{RQ2})}
Table~\ref{tab:comp_results} presents a performance comparison across all datasets for rumor verification. Our model results are averaged over 10 experimental runs, yielding a standard deviation of 0.005–0.01, with statistical significance confirmed by McNemar’s test ($p < 0.05$)~\cite{mcnemar1947note}.

We first examine models that do not explicitly incorporate stance information (listed above the midline in the table). Among these, context-aware and structure-based models such as KGs and SAMGAT outperform traditional baselines (e.g., eventAI and HT), underscoring the importance of discourse structure. Our variant without stance modeling (\textit{ours w/o stance}), which removes stance-aware encoding (stance aggregation and stance distribution) but retains our architectural backbone, outperforms all other non-stance models across datasets, highlighting the strength of our framework.

Stance-aware models (below midline) consistently outperform their non-stance counterparts. For instance, FSNet Macro-F1 improves from 0.650 to 0.710, 0.579 to 0.615, and 0.364 to 0.466 on RumEval2017, RumEval2019, and PHEME, respectively, when stance information is incorporated. Similar trends are observed for HT vs. CHT and our model with vs w/o stance. These findings validate that stance integration enhances feature discriminability for rumor verification.

Our proposed model achieves the highest Macro-F1 across all benchmarks, exceeding the best baseline (S-CoM) by 3.7\%, 0.9\%, and 4.5\% on RumEval2017, RumEval2019, and PHEME, respectively. These improvements reflect the efficacy of explicitly encoding stance signals in the semantic encoder and modeling stance cardinalities and hierarchical depth through structural covariates.

\begin{table}[t]  
\centering
\caption{Performance comparison on all datasets}  
%\resizebox{0.9\textwidth}{!}{
\resizebox{1.0\linewidth}{!}{%  
\begin{tabular}{lcccccccccc}  
\hline  
\textbf{Model} & & \multicolumn{2}{c}{RumEval2017} & & \multicolumn{2}{c}{RumEval2019} & & \multicolumn{2}{c}{PHEME} \\  
\cline{3-4} \cline{6-7} \cline{9-10}  
                                 &   & Macro-F1 & Acc & & Macro-F1 & Acc & & Macro-F1 & Acc \\  
\hline  
eventAI                          &   & 0.618        & 0.629   & & 0.577        & 0.591   & & 0.342        & 0.357   \\

HT         &   & 0.657        & 0.643   & & 0.568        & 0.572   & & 0.375        & 0.454   \\  
FSNet w/o stance                 &   & 0.650        & 0.643   & & 0.579        & 0.577   & & 0.364        & 0.576    \\  
KGs                 &   & 0.661        & 0.672   & & 0.584        & 0.593   & & 0.489        & 0.523    \\  
SAMGAT                           &   & 0.674        & 0.683   & & 0.542        & 0.562   & & 0.409        & 0.418   \\ 
Ours w/o stance                  &   & 0.681        & 0.689   & & 0.612        & 0.629   & & 0.446        & 0.557    \\
\hline

CHT  &   & 0.680        & 0.678   & & 0.579        & 0.611   & & 0.396        & 0.466   \\  

MTL-SMI                           &   & 0.685        & 0.679   & & 0.582        & 0.589   & & 0.409         & 0.468    \\ 

JRSM      &   & 0.704        & 0.717   & & 0.598        & 0.623   & & 0.448        & 0.479   \\  

FSNet with stance                 &   & 0.710        & 0.720   & & 0.615        & 0.627   & & 0.466      & 0.678    \\ 
S-CoM       &   & 0.724        & 0.723   & & 0.636        & 0.648      & & 0.621       &  0.623 \\ 
\textbf{Ours with stance }                    &   & \textbf{0.761}        &\textbf{0.768} & & \textbf{0.726 }       & \textbf{0.734}      & &\textbf{0.666}        & \textbf{0.738 }  \\ 

\hline  
\end{tabular}%  
}  

\label{tab:comp_results}  
\end{table}  

\subsubsection{\textbf{Ablation Study} (\textbf{RQ3})}
To assess the impact of each component on model performance, we conduct ablation studies on RumEval2017 and RumEval2019, as summarized in Table~\ref{tab:results_ablation}. We first evaluate a variant \textit{-stance injection}, which omits stance signal integration in the semantic encoder. Next, we replace the stance aggregator with two alternatives: (1) encoding the full conversation using BERT within its sequence length limit (\textit{-stance aggregator1}), and (2) segmenting threads into groups, encoding each with BERT, and applying MHA (\textit{-stance aggregator2}). We further ablate structural modeling through \textit{-struct cov integrator} (removing both stance distribution and hierarchical depth), \textit{-stance distribution}, \textit{-depth encoding}, and \textit{-MHA}. Lastly, we examine the effects of supplemental data preprocessing (\textit{-data prep}) and loss function choice by substituting FL with cross-entropy (\textit{-focal loss}).

\textbf{Stance-aware conversation structure modeling is crucial for debunking rumors}. As shown in the table, removing both structural covariates yields the most significant drop in performance across both datasets compared with removing either stance distribution  or hierarchical depth encoding, indicating the crucial role of combining them in modeling conversation context. On the same note, performance further declines when the structural covariates are not attended to with the MHA.  Similarly, removing stance injection also results in substantial degradation, underscoring the utility of explicit stance signal incorporation during semantic encoding. 

The reduced performance of \textit{-stance aggregator1} and \textit{-stance aggregator2} confirms that stance-wise aggregation of reply embeddings is more effective than attention-based grouping or encoding full threads constrained by BERT sequence length. Lastly, omitting supplemental preprocessing and FL yields minor declines, suggesting that handling class imbalance and text normalization, though subtle, contribute to model robustness. Overall, the complete model consistently outperforms all ablation variants across both datasets.

\begin{table}[t]  
\centering 
\caption{Ablation study on RumEval2017 and RumEval2019}  
%\resizebox{\linewidth}{!}{% 
\resizebox{0.85\linewidth}{!}{%  
\begin{tabular}{lccccccc}  
\hline  
\textbf{Model} &  & \multicolumn{2}{c}{RumEval2017} & & \multicolumn{2}{c}{RumEval2019} \\  
 \cline{3-4} \cline{6-7}  
                 &   & Macro-F1 & Acc & & Macro-F1 & Acc \\  
\hline  
-struct cov integrator &   & 0.664        & 0.672    & & 0.622         & 0.628  \\
-stance injection        &   & 0.667        & 0.671    & & 0.629         & 0.633  \\
-stance aggregator1      &   & 0.681        & 0.689    & & 0.637         & 0.639   \\
-stance aggregator2      &   & 0.686        & 0.692    & & 0.641         & 0.630   \\
-stance distribution     &   & 0.691        & 0.701    & & 0.662         & 0.675   \\  
-depth encoding    &   & 0.694        & 0.705    & & 0.671         & 0.679  \\  
-MHA                     &   & 0.712        & 0.710    & & 0.692         & 0.699   \\  
-data prep               &   & 0.717        & 0.718    & & 0.701         & 0.710 \\ 
-focal loss              &   & 0.719        & 0.728    & & 0.702         & 0.712 \\ 
\textbf{Whole model}     &  & \textbf{0.761 }       & \textbf{0.768}      & &\textbf{0.726}        & \textbf{0.734}    \\ 

\hline  
\end{tabular}%  
}  

\label{tab:results_ablation}  
\end{table} 

\subsubsection{\textbf{Early Detection} (\textbf{RQ4})}
Timely detection of rumors can mitigate their diffusion~\cite{ducci2020cascade}. We define detection checkpoints based on the elapsed time, spanning 24 hours, since the rumor post was published. At each checkpoint, only replies accumulated up to that point are considered for model evaluation.  

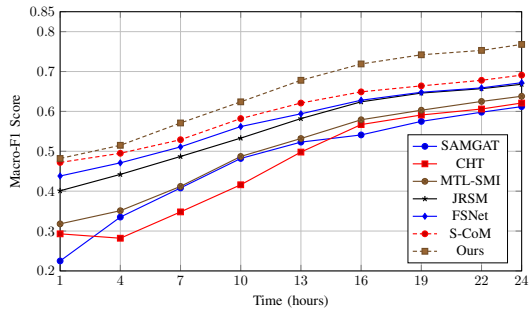
\begin{figure}[b]
    \centering
    %\resizebox{0.48\textwidth}{!}{ 
    \resizebox{0.80\linewidth}{!}{% 
    \begin{tikzpicture}
        \begin{axis}[
            xlabel={Time (hours)},
            ylabel={Macro-F1 Score},
            xmin=1, xmax=24,
            ymin=0.2, ymax=0.85,
            xtick={1,4,7,10,13,16,19,22,24},
            ytick={0.2,0.3,0.4,0.5,0.6,0.7,0.8,0.85},
            legend pos=south east,
            grid=major,
            width=13cm,
            height=8cm,
            every axis plot/.append style={thick}
        ]

        %\addplot coordinates {(1,0.195) (4,0.212) (7,0.268) (10,0.317) (13,0.389) (16,0.521) (19,0.553) (22,0.567) (24,0.579)};
        %\addlegendentry{eventAI}

        %\addplot coordinates {(1,0.184) (4,0.201) (7,0.243) (10,0.317) (13,0.372) (16,0.534) (19,0.569) (22,0.584) (24,0.601)};
        %\addlegendentry{HT}

        %\addplot coordinates {(1,0.189) (4,0.229) (7,0.289) (10,0.346) (13,0.398) (16,0.517) (19,0.558) (22,0.591) (24,0.608)};
        %\addlegendentry{KGs}

        \addplot coordinates {(1,0.225) (4,0.335) (7,0.408) (10,0.482) (13,0.523) (16,0.541) (19,0.575) (22,0.598) (24,0.612)};
        \addlegendentry{SAMGAT}

        %\addplot coordinates {(1,0.402) (4,0.442) (7,0.502) (10,0.531) (13,0.559) (16,0.572) (19,0.601) (22,0.635) (24,0.648)};
        %\addlegendentry{Ours w/o stance}

        %\addplot coordinates {(1,0.378) (4,0.417) (7,0.489) (10,0.514) (13,0.543) (16,0.567) (19,0.589) (22,0.604) (24,0.621)};
        %\addlegendentry{FSNet w/o stance}

        \addplot coordinates {(1,0.293) (4,0.282) (7,0.348) (10,0.416) (13,0.498) (16,0.567) (19,0.591) (22,0.606) (24,0.621)};
        \addlegendentry{CHT}

        \addplot coordinates {(1,0.318) (4,0.351) (7,0.412) (10,0.487) (13,0.532) (16,0.579) (19,0.603) (22,0.625) (24,0.638)};
        \addlegendentry{MTL-SMI}

        \addplot coordinates {(1,0.401) (4,0.442) (7,0.487) (10,0.533) (13,0.582) (16,0.624) (19,0.646) (22,0.657) (24,0.668)};
        \addlegendentry{JRSM}

        \addplot coordinates {(1,0.438) (4,0.471) (7,0.511) (10,0.562) (13,0.594) (16,0.628) (19,0.648) (22,0.659) (24,0.672)};
        \addlegendentry{FSNet}

        \addplot coordinates {(1,0.472) (4,0.495) (7,0.529) (10,0.582) (13,0.621) (16,0.649) (19,0.664) (22,0.678) (24,0.691)};
        \addlegendentry{S-CoM}

        \addplot coordinates {(1,0.482) (4,0.515) (7,0.571) (10,0.624) (13,0.678) (16,0.719) (19,0.742) (22,0.753) (24,0.768)};
        \addlegendentry{Ours}

        \end{axis}
    \end{tikzpicture}
    }
    \caption{Early rumor detection performance on RumEval2017 across different detection checkpoints, comparing our model against best baselines.}
    \label{fig:early-detection}
\end{figure}

Fig.~\ref{fig:early-detection} illustrates early rumor detection results on RumEval2017. The figure illustrates that all models improve as more temporal context becomes available and exhibit marginal improvement as the detection checkpoint reaches 20 hours. However, the performance gap between models becomes more pronounced in earlier stages. Notably, our proposed model outperforms all baselines from as early as 1 hour, demonstrating its effectiveness in low-resource temporal settings and value in real-world misinformation scenarios where timely intervention is crucial.

\subsubsection{\textbf{Cross-Platform Adaptation} (\textbf{RQ5})}

For an evaluation of the generalizability of the model across social media platforms, we capitalize on the composition of the RumEval2019 dataset of Twitter and Reddit posts. We partition the dataset by platform and conduct two sets of experiments: (1) training on one platform and testing on the other to evaluate cross-platform transferability; and (2) training and testing on the same platform to establish in-domain performance baselines. Results are presented in Table~\ref{tab:cross_platform}, which indicates that the model achieves the highest performance when trained and tested on the same platform. When applied in a cross-platform setting, the model maintains competitive performance, with only negligible degradation. These results indicate that while platform-specific training yields optimal results, the model exhibits strong cross-platform generalization capabilities, making it adaptable for deployment across diverse social media domains.
\begin{table}[t]
\centering
\caption{Cross-Platform Adaptation Results on RumEval2019}
\resizebox{0.70\linewidth}{!}{
\begin{tabular}{lcccc}
\hline
\textbf{Train Platform} & \textbf{Test Platform} & \textbf{Macro-F1} & \textbf{Accuracy} \\
\hline
Twitter     & Twitter     & 0.761 & 0.768 \\
Reddit      & Reddit      & 0.721 & 0.714 \\
Twitter     & Reddit      & 0.701 & 0.703 \\
Reddit      & Twitter     & 0.696 & 0.708 \\
\hline
\end{tabular}
}
\label{tab:cross_platform}
\end{table}

\section{Conclusion}\label{sec:conclusion}
This study introduces a stance-aware structural modeling framework for rumor verification that jointly encodes the semantic content and stance signals of social media posts. To overcome sequence 
 length constraints inherent in transformer-based models, we propose a stance-based reply aggregation strategy that compresses discourse into fixed-size representations by stance category. We further enhance the model with two structural covariates, stance distribution and hierarchical depth, that capture stance imbalance and the structural influence of replies within a controversy. Experimental results on benchmark datasets demonstrate that our model substantially improves rumor verification performance, particularly in early detection scenarios and cross-platform settings. Ablation studies validate the effectiveness of incorporating both stance and structural signals, while comparative analyses confirm that explicitly modeling stance yields more discriminative features.  Despite the cost of manual stance annotation, our findings reaffirm the critical role of stance information in rumor verification. Future work will explore generating pseudo stance labels via weak and self-supervised methods, extend the framework to include visual stance cues for multi-modal rumor verification, and evaluate the model on diverse and multilingual datasets.

\textit{Ethical statement:} while this study has positive implications, ethical challenges and risks persist. These include false classifications that may suppress truthful content or spread falsehoods, and biases in training data that could lead to unfair outcomes. Mitigation strategies such as human validation, bias auditing, and fairness checks are essential. The system’s potential misuse for censorship highlights the need for transparency and strict ethical oversight. Publicly available anonymized datasets  are used, but researchers must prevent the amplification of harmful content and adhere to ethical standards and platform policies.

\section*{ACKNOWLEDGMENT}
This work is supported by the National Science Foundation (NSF) under Award number OIA-1946391, Data Analytics that are Robust and Trusted (DART). 

%\section*{References}

%\printbibliography
%\IEEEtriggeratref{13}
%\sloppy
\bibliographystyle{IEEEtran}
\bibliography{conference_101719}

\end{document}